\documentclass[10pt,twocolumn,letterpaper]{article}

\usepackage[pagenumbers]{cvpr} 

\usepackage{graphicx}
\usepackage{amsmath}
\usepackage{amssymb}
\usepackage{booktabs}
\usepackage{multirow}
\usepackage{makecell}

\usepackage[pagebackref,breaklinks,colorlinks]{hyperref}

\usepackage[capitalize]{cleveref}
\crefname{section}{Sec.}{Secs.}
\Crefname{section}{Section}{Sections}
\Crefname{table}{Table}{Tables}
\crefname{table}{Tab.}{Tabs.}

\def\cvprPaperID{3500} 
\def\confName{CVPR}
\def\confYear{2022}

\begin{document}

\title{Non-linear Motion Estimation for Video Frame Interpolation using Space-time Convolutions}

\author{Saikat Dutta \qquad Arulkumar Subramaniam \qquad Anurag Mittal\\
Indian Institute of Technology Madras\\
Chennai, India\\
{\tt\small \{cs18s016,aruls,amittal\}@cse.iitm.ac.in }
}
\maketitle

\begin{abstract}
Video frame interpolation aims to synthesize one or multiple frames between two consecutive frames in a video. It has a wide range of applications including slow-motion video generation, frame-rate up-scaling and developing video codecs. Some older works tackled this problem by assuming per-pixel linear motion between video frames. However, objects often follow a non-linear motion pattern in the real domain and some recent methods attempt to model per-pixel motion by non-linear models (e.g., quadratic). A quadratic model can also be inaccurate, especially in the case of motion discontinuities over time (i.e. sudden jerks) and occlusions, where some of the flow information may be invalid or inaccurate.

In our paper, we propose to approximate the per-pixel motion using a space-time convolution network that is able to adaptively select the motion model to be used. Specifically, we are able to softly switch between a linear and a quadratic model.
Towards this end, we use an end-to-end 3D CNN encoder-decoder architecture over bidirectional optical flows and occlusion maps to estimate the non-linear motion model of each pixel. 
Further, a motion refinement module is employed to refine the non-linear motion and the interpolated frames are estimated by a simple warping of the neighboring frames with the estimated per-pixel motion. Through a set of comprehensive experiments, we validate the effectiveness of our model and show that our method outperforms state-of-the-art algorithms on four datasets (Vimeo, DAVIS, HD and GoPro).
\end{abstract}

\vspace{-0.5cm}
\section{Introduction}
\label{sec:intro}

Video frame interpolation (VFI) (also known as Video temporal super-resolution) is a significant video enhancement problem which aims to synthesize one or more visually coherent frames between two consecutive frames in a video, i.e., to up-scale the number of video frames. Such an up-scaling method finds its usage in numerous video-based applications such as slow-motion video generation (e.g., in sports and TV commercials), video compression-decompression framework \cite{bframe}, generating short videos from GIF images \cite{gif2vid}, novel view synthesis \cite{flynn2016deepstereo} and medical imaging \cite{karargyris2010three,zinger2011view}. 

Earlier methods \cite{dvf,superslomo,toflow,featureflow,park2020bmbc} in this domain rely on estimating optical flow between interpolated frame to source frames (i.e., neighboring frames). Once the optical flow is estimated, the interpolated frame can be synthesized by a simple warp operation from source images. However, estimating an accurate optical flow between video frames is a hard problem in itself. Thus, some methods \cite{adaconv, sepconv, lee2019learning} relied on estimating per-pixel interpolation kernels to smoothly blend source frames to produce the interpolated frame. Further, some hybrid methods  \cite{memc,dain} were also proposed to integrate optical flow and interpolation kernel based approaches, exhibiting better performance than earlier class of methods.

Most state-of-the-art interpolation algorithms take two neighboring frames as input to produce the intermediate frame. As a result, only a linear motion can be modeled between the frames either explicitly or implicitly. However, objects often follow complex, non-linear trajectories. To this end, researchers recently focused on leveraging information from more than two neighboring frames \cite{xu2019quadratic, mprn, all_at_once, tridirectional, kalluri2020flavr}.

3D convolutional neural networks have gained success in many important computer vision tasks such as action recognition \cite{hara2018can, ji20123d,tran2018closer, hara2017learning}, object recognition \cite{maturana2015voxnet}, video object segmentation \cite{hou2019efficient} and biomedical volumetric image segmentation \cite{3dunet}. However, the application of 3D CNNs in VFI task is largely unexplored. Recently, Kalluri et al. \cite{kalluri2020flavr} use a 3D UNet to directly synthesize interpolated frames. However, hallucinating pixel values from scratch can lead to blurry results and simply copying pixels from nearby frames can produce better results \cite{dvf}.

In this work, we propose a novel frame interpolation method. First, we compute bi-directional flow and occlusion maps from four neighboring frames and predict a non-linear flow model with the help of a 3D CNN. In this regard, we formulate a novel 3D CNN architecture namely ``GridNet-3D'' inspired from \cite{yuanchen2021gridnet} for efficient multi-scale feature aggregation. 
Further, the predicted non-linear flow model is used as coefficients in a quadratic formulation of inter-frame motion. The idea is that such an approach can adaptively select between linear and quadratic models by estimating suitable values for the coefficients. 
Intermediate backward flows are produced through flow reversal and motion refinement. Finally, two neighboring frames are warped and combined using a blending mask to synthesize the interpolated frame. Our algorithm demonstrates state-of-the-art performance over existing approaches on multiple datasets.

Our main contributions are summarized as follows:
\begin{itemize}
    \setlength\itemsep{0.2em}
    \item We introduce a novel frame interpolation algorithm that utilizes both flow and occlusion maps between four input frames to estimate an automatically adaptable pixel-wise non-linear motion model to interpolate the frames.
    \item We propose a parameter and runtime-efficient 3D CNN named ``GridNet-3D'' to aggregate multi-scale features efficiently.
    \item Through a set of comprehensive experiments on four publicly available datasets (Vimeo, DAVIS, HD and GoPro), we demonstrate that our method achieves state-of-the-art performance.
\end{itemize}

Rest of the paper is organized as follows: Section \ref{sec:relatedworks} discusses some significant prior works in Video Frame Interpolation, Section \ref{proposed} describes our algorithm, Section \ref{expt} contains experiments along with some ablation studies and finally Section \ref{sec:conclusion} summarizes limitations of our approach and discusses possible future directions.

\section{Related work}
\label{sec:relatedworks}

In this section, we briefly describe the methods that are relevant to this paper.

\vspace{-0.3cm}
\paragraph{Video-frame interpolation (VFI):}  Based on the type of motion cues used, VFI methods can be mainly classified into three categories: \textbf{1) Optical flow based approaches:} In this class of methods, optical flow \cite{flownet2,meister2018unflow,pwcnet} is predominantly used as a motion cue for interpolating frames. Recent state-of-the-art methods use fully convolutional networks (FCN) \cite{dvf}, 2D UNets \cite{superslomo, unet}, multi-scale architectures \cite{toflow,spynet}, or bilateral cost volume \cite{park2020bmbc} to predict backward optical flows to warp the neighboring frames and estimate the interpolated frame. In some methods, forward optical flow is also utilized \cite{ctxsyn, softsplat}.
However, these class of methods rely on estimating the optical flow based on 2D CNNs at the frame-level that might not capture motion features well. Our work belongs to this category with a key difference in that we use a 3D CNN to capture a large spatiotemporal stride to better estimate per-pixel non-linear motion. \textbf{2) Phase-based approaches:} Estimating accurate optical flow is a hard problem, especially when it involves large motion, illumination variations and motion blur. An alternative cue to optical flow is to use phase-based modification of pixels. These methods estimate low-level features such as per-pixel phase \cite{phase2015}, fourier decomposition of images using steerable pyramids \cite{phasenet}, phase and amplitude features using one-dimensional separable Gabor filters \cite{zhou2019frame} to estimate the interpolated frame. \textbf{3) Kernel-based approaches:} Different from optical flow and phase-based methods, kernel-based methods strive to estimate per-pixel kernels to blend patches from neighborhood frames. Some of these methods employ adaptive convolution \cite{adaconv}, adaptive separable convolution \cite{sepconv} or adaptive deformable convolutions \cite{lee2019learning}.

Apart from these categories, some methods proposed a hybrid approach to utilize the advantage from multiple cues. For instance, a combination of interpolation kernels and optical flow \cite{memc} or optical flow and depth \cite{dain} are employed to obtain complementary features.

\vspace{-0.3cm}
\paragraph{Multi-frame VFI approaches:} Recent methods have started using multiple frames to capture complex motion dynamics between frames. For instance, Choi et al. \cite{tridirectional} utilize three frames and bi-directional optical flow between them to generate the intermediate flows and use warping and frame generation module to estimate the final interpolated frame. Chi et al. \cite{all_at_once} use cubic modeling and a pyramid style network to produce seven intermediate frames. Similarly, Xu et al. \cite{xu2019quadratic} use four frames to model a quadratic motion between frames. They estimate quadratic motion parameters in terms of an analytical solution involving optical flow.
Our method uses four frames and estimates non-linear (quadratic) motion model similar to \cite{xu2019quadratic}. However, we show that using a powerful 3D CNN to estimate the motion parameters instead of an analytical solution significantly performs better (ref. Section \ref{sec:sota_comparison}).

\vspace{-0.3cm}
\paragraph{3D CNN models:} 3D CNNs are prevalent in Computer Vision tasks involving a spatio-temporal input (video-based tasks) such as action recognition \cite{hara2018can, ji20123d,tran2018closer, hara2017learning}, video object segmentation \cite{hou2019efficient} and video captioning \cite{chen2019temporal,aafaq2019spatio}. Related to VFI task, Zhang et al. \cite{mprn} developed a Multi-frame Pyramid Refinement (MPR) scheme using 3D UNet to estimate intermediate flow maps from four input frames. Kalluri et al. \cite{kalluri2020flavr} utilize a 3D encoder-decoder architecture to directly synthesize interpolated frames from four input frames. Differing from these methods, we use 3D CNN over optical flow and occlusion maps to predict non-linear motion coefficients.

\begin{figure*}[!ht]
    
    \centering
    \includegraphics[width=0.9\textwidth]{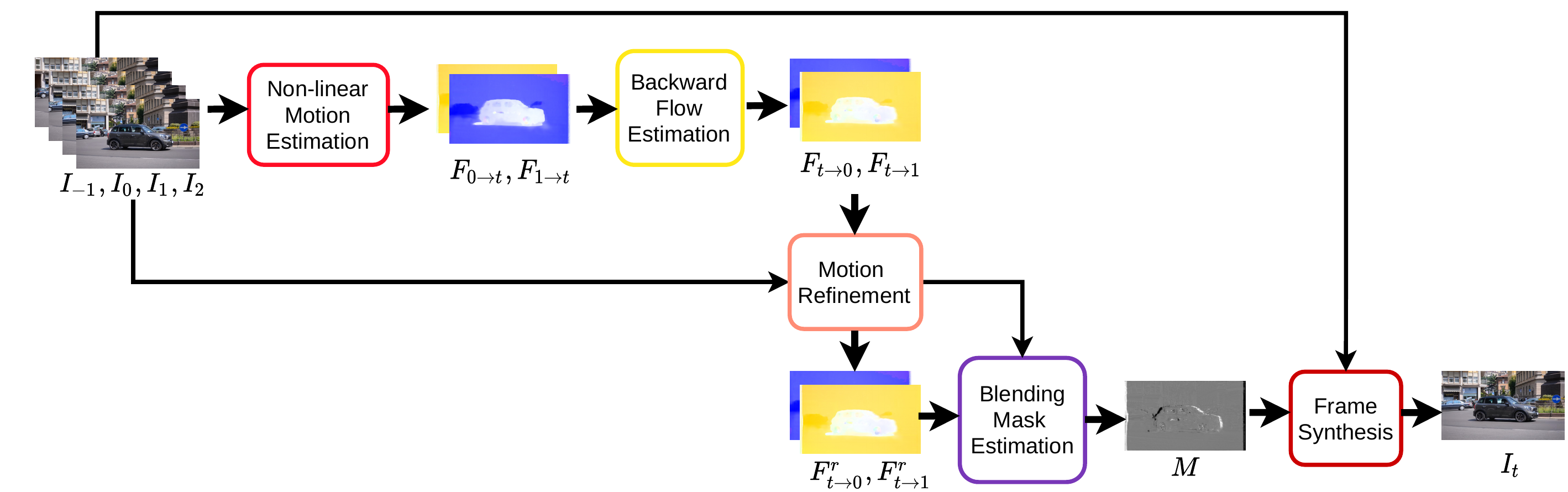}
    \vspace{-0.3cm}
    \caption{Overview of our interpolation algorithm. Non-linear motion estimation module produces forward flow ($ F_{0\rightarrow t}, F_{1\rightarrow t}$), which is used to generate backward flow ($F_{t\rightarrow 0},F_{t\rightarrow 1}$). This backward flow is refined using a motion refinement module. Finally, a blending mask $M$ is estimated that is used to fuse the warped frames to generate interpolated frame $I_t$.}
    \label{overview}
    \vspace{-0.3cm}
\end{figure*}

\begin{figure*}[!ht]
    
    \centering
    \includegraphics[width=0.9\textwidth]{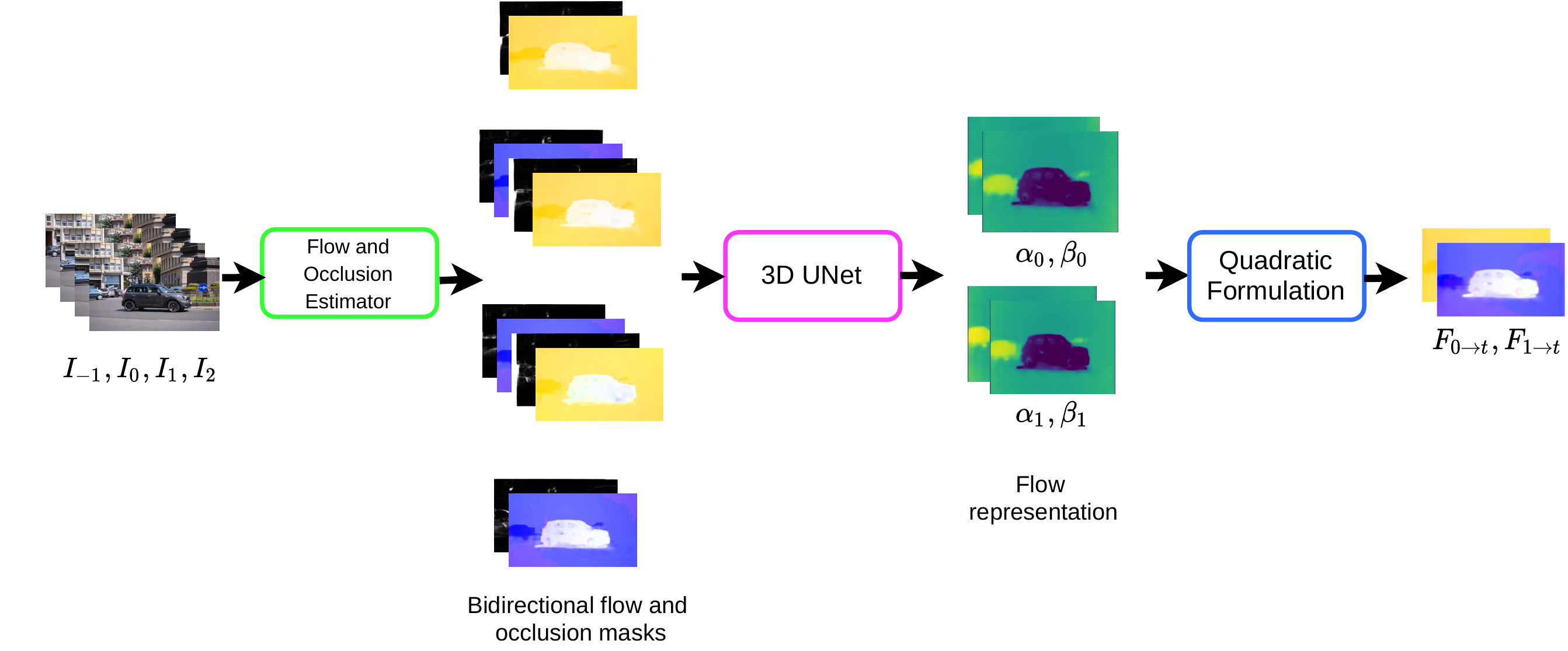}
    \vspace{-0.3cm}
    \caption{Non-linear motion estimation module. First, bi-directional flow and occlusion maps are estimated which are fed to a 3D UNet to generate flow representation $\alpha, \beta$. This flow representation is used to produce forward intermediate flow using quadratic formulation.}
    \label{NME}
    \vspace{-0.5cm}
\end{figure*}

\section{Space-time convolution network for non-linear motion estimation}
\label{proposed}

Determining the motion trajectory of pixels is essential to determine the transition of pixel values from one frame to the next.  Traditional methods use optical flow to achieve this goal with the assumption of brightness constancy and velocity smoothness constraint and
use a linear model for interpolation. 
While some methods recently have used a quadratic model for flow estimation with improved results, such a model is not applicable in certain scenarios such as motion discontinuities and occlusions.
In this work, we opt to use a 3D CNN encoder-decoder architecture to estimate per-pixel non-linear motion that can easily switch between a linear and quadratic model.  Specifically, the 3D CNN takes  a set of bi-directional optical flows and occlusion maps between consecutive video frames \{$I_{-1}, I_0, I_1, I_2$\} to estimate the non-linear motion model that is utilized by other modules to predict an interpolated frame $I_t$, where $t \in (0,1)$. i.e., the output frame $I_t$ needs to be coherent in terms of appearance and motion between $I_0$ and $I_1$.

An overview of our framework is shown in Figure \ref{overview}. The framework consists of the five modules namely: 1) Non-linear motion estimation (NME) module, 2) Backward flow estimation (BFE) module, 3) Motion refinement (MR) module, 4) Blending mask estimation (BME) module, and 5) Frame synthesis.  The details of each module are described in the following sections.

\subsection{Non-linear motion estimation (NME) module}
\label{sec:nme_module}

\textcolor{blue}{}

Recent methods attempt to overcome linear motion assumption by modeling a non-linear motion. Xu et al.       \cite{xu2019quadratic} proposed to model a quadratic motion model in terms of time $t$. i.e., with an assumption that pixel motion follows a quadratic motion of form $\alpha t + \beta t^2$. They estimate the motion model parameters $\alpha, \beta$ by an analytical formula derived using per-pixel optical flow. 
However, such a quadratic assumption cannot be applied to the pixels involving unreliable optical flow estimates (\eg occluded pixels). Using such unreliable optical flow estimates may lead to inaccurate intermediate flow estimation and may end up with erroneous interpolation results. Instead of directly estimating quadratic motion parameters from optical flow, we attempt to estimate $\alpha, \beta$ through a 3D CNN model. 

To learn suitable $\alpha$ and $\beta$ in the non-linear motion model, given the input frames \{$I_{-1},I_0, I_1, I_2$\}, we first estimate bi-directional flow and occlusion maps between neighboring frames using a pre-trained \textit{PWCNet-Bi-Occ} network \cite{irr}.  Architecture-wise, \textit{PWCNet-Bi-Occ} is based on state-of-the-art optical flow network \textit{PWCNet} \cite{pwcnet}.  It takes two frames \{$I_x, I_{y}$\} as input and extracts multi-scale feature maps of each frame.  At each scale, a correlation volume is computed between corresponding feature maps of the frames.  Then, bidirectional optical flows \{$F_{x\rightarrow{y}}, F_{y\rightarrow{x}}$\} and occlusion maps \{$O_{x\rightarrow{y}}, O_{y\rightarrow{x}}$\} are obtained as output at each level by following a coarse-to-fine strategy.  We use the optical flow and occlusion map outputs from the finer level in our work.

The bi-directional optical flows $\{F_{i\rightarrow{(i+1)}}, F_{{(i+1)}\rightarrow i}\}_{i=-1}^2$ and occlusion maps $\{O_{i\rightarrow{(i+1)}}, O_{{(i+1)}\rightarrow i}\}_{i=-1}^2$ are arranged in temporal order and results in a 5D tensor of size $B \times  6 \times \text{\#frames} \times H \times W$.  
Here $B, H, W$ denote batch size, height and width respectively, and the 6 channels belong to bi-directional optical flows 
and occlusion maps.  
This tensor is passed through a 3D CNN model to estimate a representation of dimension $B\times 4 \times 2 \times H \times W$.  The temporal dimension of 2 corresponds to $t=0$ and $t=1$. In each temporal slice, we predict two coefficient maps $\alpha$ and $\beta$, each of 2-dimensions.  We refer these coefficients $\alpha, \beta$ as the flow representation. Now the per-pixel non-linear motion $F_{0\rightarrow t}$ of frame $I_0$ towards the interpolated frame $I_t$ is given by:
\begin{equation}
    F_{0\rightarrow t} = \alpha_0 \times t + \beta_0 \times t^{2}
\end{equation}    

Similarly, $F_{1\rightarrow t}$ is given by:
\begin{equation}
    F_{1\rightarrow t} = \alpha_1 \times (1-t) + \beta_1 \times (1-t)^{2}
\end{equation}  

Estimating the coefficients $\alpha_0, \beta_0, \alpha_1$ and $\beta_1$ through a neural network instead of an analytical solution \cite{xu2019quadratic} offers the following advantages:  1) The network can flexibly choose between linear and non-linear motion. For pixels to follow a linear motion, the network may predict $\beta = 0$; 2) Unlike \cite{xu2019quadratic}, learned estimates of $\alpha$'s, $\beta$'s are better equipped to handle occlusion by utilizing the occlusion maps, 3) Having access to large temporal receptive field of 4 frames, the non-linear motion coefficients estimated through a 3D CNN can determine more accurate motion than \cite{xu2019quadratic} which rely on optical flow to estimate the coefficients. Figure \ref{NME} shows the pipeline of the non-linear motion estimation module.

\begin{figure}
    \centering
    \includegraphics[width=0.5\textwidth,keepaspectratio]{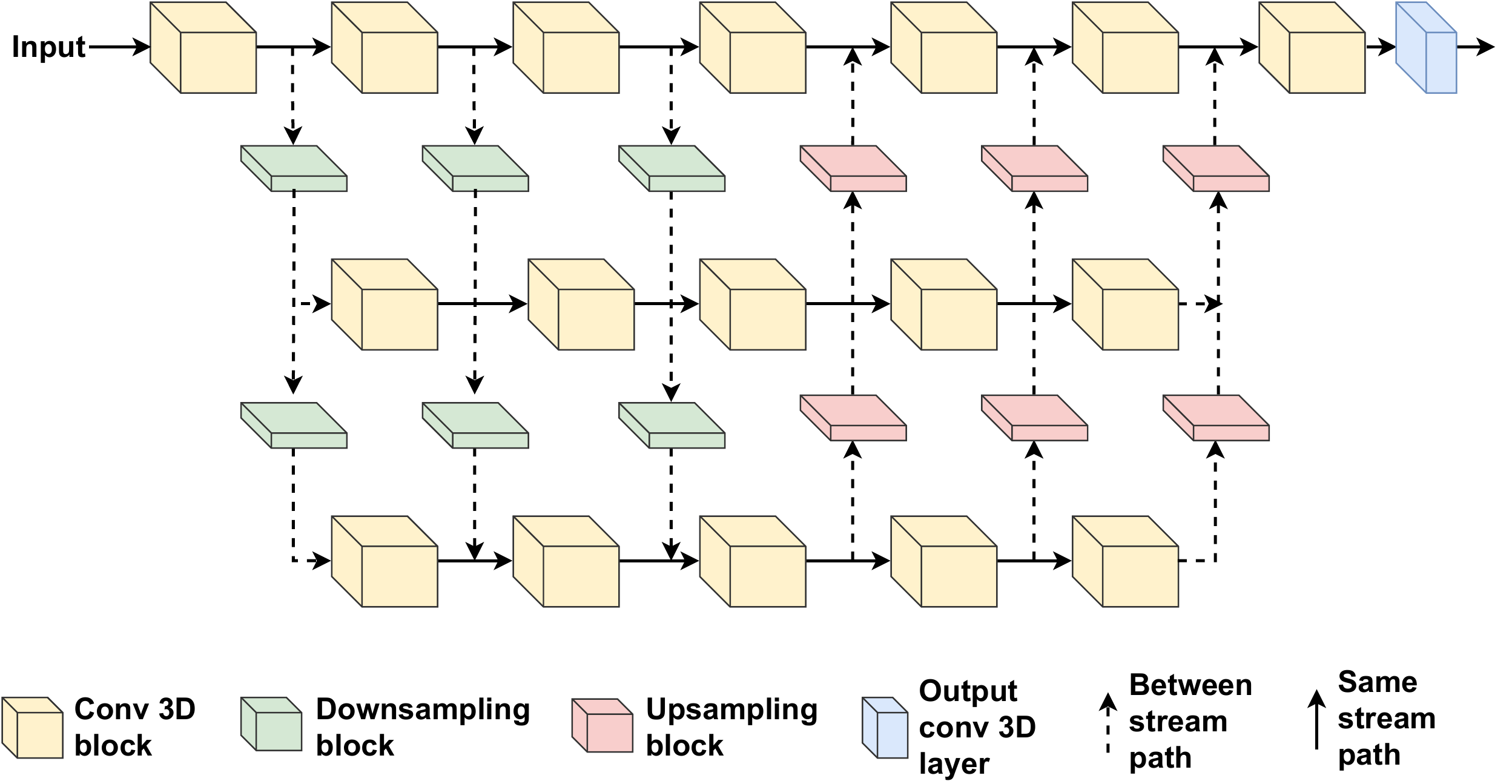}
    \vspace{-0.3cm}
    \caption{Novel GridNet-3D architecture for efficient multi-scale feature aggregation inspired from \cite{yuanchen2021gridnet}. It consists of three parallel streams operating in different feature resolutions and the communication between streams is handled by \textit{downsampling} and \textit{upsampling} blocks (refer Sec. \ref{sec:nme_module}).}
    \vspace{-0.7cm}
    \label{3dgridnet}
\end{figure}

\textbf{Network specification:} We formulate the NME module to predict $\alpha, \beta$ with two crucial design choices in mind: 1) to capture spatiotemporal features and 2) to incorporate multi-scale features efficiently. 
3D CNN networks are the natural choices to capture spatiotemporal features among video frames. 
However, the existing architectures for pixel-wise tasks (\textit{e.g.}, UNet-3D \cite{kalluri2020flavr}) adopt a single-stream Encoder-Decoder style architecture that aggregates multi-scale features by the process of sequential downsampling and skip-connection which may result in information loss \cite{fourure2017residual}.  
Inspired by the success of GridNet \cite{gridnet,ctxsyn} in efficiently incorporating multi-resolution features, we formulate a novel 3D version of GridNet namely ``GridNet-3D'' by replacing its 2D convolutional filters with 3D convolutional filters. 
\textit{GridNet-3D} consists of three parallel streams to capture features with different resolutions and each stream has five convolutional blocks arranged in a sequence as shown in Fig. \ref{3dgridnet}. Each convolutional block is made up of two conv-3D layers with a residual connection. The three parallel streams have channel dimensions of 16, 32 and 64 respectively. 
The communication between the streams are handled by a set of \textit{downsampling} and \textit{upsampling} blocks. The \textit{downsampling} block consists of spatial max pooling of stride 2 followed by one conv-3D layer, whereas the \textit{upsampling} block consists of one bilinear upsampling layer followed by two conv-3D layers. 

We perform a set of comparative studies with prevalent architectures such as UNet-3D \cite{kalluri2020flavr}, UNet-2D \cite{superslomo} and demonstrate the results in the Sec. \ref{modelconfig}.

\subsection{Backward flow estimation (BFE) module}
The non-linear motions ($F_{0\rightarrow t}, F_{1\rightarrow t}$) estimated in NME module are forward intermediate flows.
To make use of backward warping operation \cite{stn} on the frames $I_0$ and $I_1$, we require the backward intermediate flows ($F_{t \rightarrow 0}$ and $F_{t \rightarrow 1}$) to be determined. To achieve this, we use a differentiable flow reversal layer proposed by \cite{xu2019quadratic} to obtain $F_{t \rightarrow 0}$ and $F_{t \rightarrow 1}$ from $F_{0 \rightarrow t}$ and $F_{1 \rightarrow t}$ respectively. Backward flow at a pixel position $\textbf{x}$ is formulated as weighted average of forward flows of all pixels $\textbf{p}$ that fall into neighborhood of pixel $\textbf{x}$. $F_{t \rightarrow 0}$ at pixel position $\textbf{x} = (x,y)$ is given by, 
\small{
\begin{equation}
F_{t \rightarrow 0}(\textbf{x}) = \frac{ \sum_{\textbf{p}+  F_{0 \rightarrow t}(\textbf{p}) \in N(\textbf{x})} w(\textbf{x}, \textbf{p} + F_{0 \rightarrow t}(\textbf{p}))(-F_{0 \rightarrow t}(\textbf{p}))}{\sum_{\textbf{p}+  F_{0 \rightarrow t}(\textbf{p} + F_{0 \rightarrow t}(\textbf{p})) \in N(\textbf{x})} w(\textbf{x}, \textbf{p})}
\label{flow_reversal}
\end{equation}
}
where $N(\textbf{x})$ denotes a $2\times2$ neighborhood around $\textbf{x}$ and $w(.,.)$ is a weighting function given by:
\begin{equation}
    w(\textbf{a},\textbf{b}) = e^{-||\textbf{a} -\textbf{b}||^2_2}
\end{equation}

Following similar procedure in Equation \ref{flow_reversal}, $F_{t\rightarrow 1}$ is computed from $F_{1 \rightarrow t}$.

\subsection{Motion refinement (MR) module}

To further refine the estimated backward flows ($F_{t \rightarrow 0}$ and $F_{t \rightarrow 1}$), we use a learning based motion refinement approach \cite{xu2019quadratic}. To this end, the refinement network takes concatenated source frames, warped frames and flow maps as input and applies a fully convolutional network to generate per-pixel offset $(\Delta x, \Delta y)$ and residuals ($r(x, y)$).

Refined optical flow, $F^r_{t\rightarrow 0}$ at pixel $(x,y)$ is given by: 
\begin{equation}
    F^r_{t\rightarrow 0}(x,y) = F_{t\rightarrow 0}(x+\Delta x, y+\Delta y) + r(x,y)
\end{equation}

$F_{t\rightarrow 1}$ is refined in a similar manner to obtain $F^r_{t\rightarrow 1}$.

We try with two types of motion refinement network in this work namely: 1) UNet-2D \cite{unet,xu2019quadratic}, and 2) GridNet-2D \cite{gridnet, ctxsyn, dutta2021efficient}. Finally, we choose GridNet-2D as the motion refinement network due to its superior performance (ref. Section \ref{ablation_sec}).

\subsection{Blending mask estimation (BME) module}
The refined backward motions $F^r_{t \rightarrow 0}$ and $F^r_{t \rightarrow 1}$ are used to warp images $I_0$ and $I_1$ to yield two estimates $I_{t0}, I_{t1}$ for interpolated frame $I_t$. However, merging these two estimates is not straight-forward. The naive approach of averaging the two estimates and using it as interpolated frame $I_t$ gives sub-par results. To improve the quality of interpolated frame, we use a learnable CNN that takes input as the stack of warped frames and intermediate feature maps from previous step to output a soft blending mask $M$. The BME module consists of three convolutional layers followed by a sigmoid activation function \cite{xu2019quadratic} to generate the mask $M$.

\subsection{Frame synthesis} 
We linearly blend the warped frame using blending mask \cite{superslomo} computed from the BME module. The final interpolated frame $I_t$ is given by:
\small{
\begin{equation}
 \hat{I}_{t} = \frac{ (1-t) \times M \odot bw(I_0, F^r_{t\rightarrow0}) + t \times (1-M) \odot bw(I_1, F^r_{t\rightarrow1})}  
    {(1-t) \times M + t \times (1-M)}
\label{hr_syn}
\end{equation}
}
where $bw(.,.)$ denotes the backward warping function. 


\section{Datasets, Experiments, Results }
\label{expt}
\subsection{Datasets}

We have used the following datasets of different image resolutions in our experiments.

\textbf{Vimeo Septuplet dataset:} Vimeo Septuplet dataset \cite{toflow} consists of 72,436 frame-septuplets of resolution $256 \times 448$. This dataset is divided into a training subset of 64,612 septuplets and a test subset of 7,824 septuplets.  We use 1\textsuperscript{st}, 3\textsuperscript{rd}, 5\textsuperscript{th} and 7\textsuperscript{th} frame from the septuplets as input frames and predict the 4\textsuperscript{th} frame as interpolation ground truth.  We use training subset of this dataset for training and evaluate the model on other datasets without fine-tuning.

\textbf{DAVIS dataset:} DAVIS-2017 TrainVal dataset \cite{davis} contains 90 video clips with diverse scenes and complex motions.  We utilize its 480p counterpart for evaluation purposes.  We extract 2,849 quintuplets from the provided video sequences.  

\textbf{HD dataset:} Bao et al. \cite{memc} collected 11 HD videos consisting of four 544p, three 720p and four 1080p videos.  We extract 456 quintuplets from these videos and discard 8 quintuplets with blank frames and scene changes.  Finally, we use 448 quintuplets for evaluation.

\textbf{GoPro dataset:} GoPro dataset proposed by Nah et al. \cite{nah2017deep} contains 33 720p videos captured at 720 FPS.  We extract 1,500 sets of 25 images from the test split consisting of 11 videos.  We use 1\textsuperscript{st}, 9\textsuperscript{th}, 17\textsuperscript{th} and 25\textsuperscript{th} frames as input frames and 13\textsuperscript{th} frame is used as the interpolation target.

\subsection{Training Details}

We develop our models using the Pytorch \cite{pytorch} framework.
During training, we optimize the network using Adam optimizer \cite{adam} with the following hyper-parameters: batch size = 64, $\beta_1=0.9$ and $\beta_2=0.999$, input frame size = random crop of $256\times256$.  The learning rate is initially set to $2\times10^{-4}$ and is divided by a factor of 10 when the loss plateaus. The \textit{PWCNet-Bi-Occ} network \cite{irr} is fixed until the learning rate reaches the value $2\times10^{-6}$ and then, it is fine-tuned with the whole network.  The model takes around 16 epochs to converge. Code will be released in Github upon acceptance.

\begin{table*}[!ht]
\caption{Effect of different CNN architectures used in NME module. Best and second best scores are colred in \textcolor{red}{{red}} and \textcolor{blue}{{blue}} respectively.}
\vspace{-0.3cm}
\small
\centering
\begin{tabular}{|c|c|c|c|c|c|c|c|c|c|c|}
\hline
\multirow{2}{*}{\begin{tabular}[c]{@{}c@{}}CNN used \\ in NME\end{tabular}} & \multicolumn{2}{c|}{Vimeo Septuplet} & \multicolumn{2}{c|}{DAVIS}    & \multicolumn{2}{c|}{HD}       & \multicolumn{2}{c|}{GoPro}    & \multirow{2}{*}{\begin{tabular}[c]{@{}c@{}}Params\\ (M)\end{tabular}} & \multirow{2}{*}{\begin{tabular}[c]{@{}c@{}}Runtime\\ (s)\end{tabular}} \\ \cline{2-9}
                                                                            & \textbf{PSNR}     & \textbf{SSIM}    & \textbf{PSNR} & \textbf{SSIM} & \textbf{PSNR} & \textbf{SSIM} & \textbf{PSNR} & \textbf{SSIM} &                                                                         &                                                                          \\ \hline
UNet-2D                                                                     & 34.76             & 0.9537           & 27.34         & 0.8254        & 31.21         & 0.8971        & 28.90         & 0.8793        &  \textcolor{blue}{38.30}                                                                       &    \textcolor{red}{0.18}                                                                      \\ \hline
UNet-3D                                                                     & \textcolor{blue}{34.96}             & \textcolor{red}{0.9545}           & \textcolor{blue}{27.46}         & \textcolor{blue}{0.8278}        & \textcolor{blue}{31.31}         & \textcolor{blue}{0.8976}        & \textcolor{blue}{29.01}         & \textcolor{blue}{0.8826}        &     60.55                                                                    &       0.37                                                                   \\ \hline
GridNet-3D                                                                     & \textcolor{red}{34.99}             & \textcolor{blue}{0.9544}           & \textcolor{red}{27.53}         & \textcolor{red}{0.8281}        & \textcolor{red}{31.49}         & \textcolor{red}{0.9000}        & \textcolor{red}{29.08}         & \textcolor{red}{0.8826}        &   \textcolor{red}{20.92}                                                                      &        \textcolor{blue}{0.32}                                                                  \\ \hline
\end{tabular}
\label{nme_abl_tab}
\end{table*}

\begin{figure*}[!ht]
    \centering
    \includegraphics[width=0.65\textwidth]{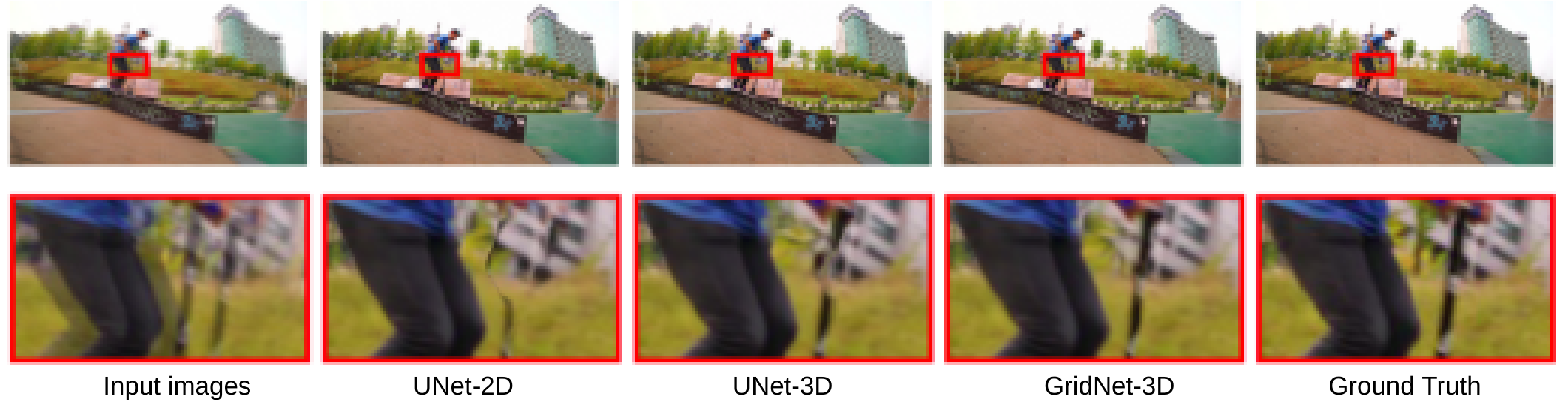}
    \vspace{-10px}
    \caption{Qualitative comparison between different CNN architectures used in NME module.}
    \label{nme_abl_fig}
\end{figure*}

\begin{table*}[!ht]
\caption{Quantitative comparison between UNet and GridNet as MR module.}
\vspace{-0.3cm}
\centering
\small
\begin{tabular}{|c|cc|cc|cc|cc|c|c|}
\hline
\multirow{2}{*}{\begin{tabular}[c]{@{}c@{}}Motion Refinement\\ module\end{tabular}} & \multicolumn{2}{c|}{Vimeo Septuplet}                  & \multicolumn{2}{c|}{DAVIS}                            & \multicolumn{2}{c|}{HD}                               & \multicolumn{2}{c|}{GoPro}                            & \multirow{2}{*}{\begin{tabular}[c]{@{}c@{}}Params\\ (M)\end{tabular}} & \multirow{2}{*}{\begin{tabular}[c]{@{}c@{}}Runtime\\ (s)\end{tabular}} \\ \cline{2-9}
                                                                                    & \multicolumn{1}{c|}{\textbf{PSNR}}  & \textbf{SSIM}   & \multicolumn{1}{c|}{\textbf{PSNR}}  & \textbf{SSIM}   & \multicolumn{1}{c|}{\textbf{PSNR}}  & \textbf{SSIM}   & \multicolumn{1}{c|}{\textbf{PSNR}}  & \textbf{SSIM}   &                                                                       &                                                                        \\ \hline
UNet-2D                                                                                & \multicolumn{1}{c|}{34.70}          & 0.9532          & \multicolumn{1}{c|}{27.32}          & 0.8260          & \multicolumn{1}{c|}{31.02}          & 0.8944          & \multicolumn{1}{c|}{28.81}          & 0.8798          & 78.11                                                                 & \textbf{0.37}                                                          \\ \hline
GridNet-2D                                                                             & \multicolumn{1}{c|}{\textbf{34.96}} & \textbf{0.9475} & \multicolumn{1}{c|}{\textbf{27.46}} & \textbf{0.8278} & \multicolumn{1}{c|}{\textbf{31.31}} & \textbf{0.8976} & \multicolumn{1}{c|}{\textbf{29.01}} & \textbf{0.8826} & \textbf{60.55}                                                        & \textbf{0.37}                                                          \\ \hline
\end{tabular}
\centering
\label{flow_ref}
\vspace{-0.3cm}
\end{table*}

\begin{figure*}[!ht]
    \centering
    \includegraphics[width=0.7\textwidth]{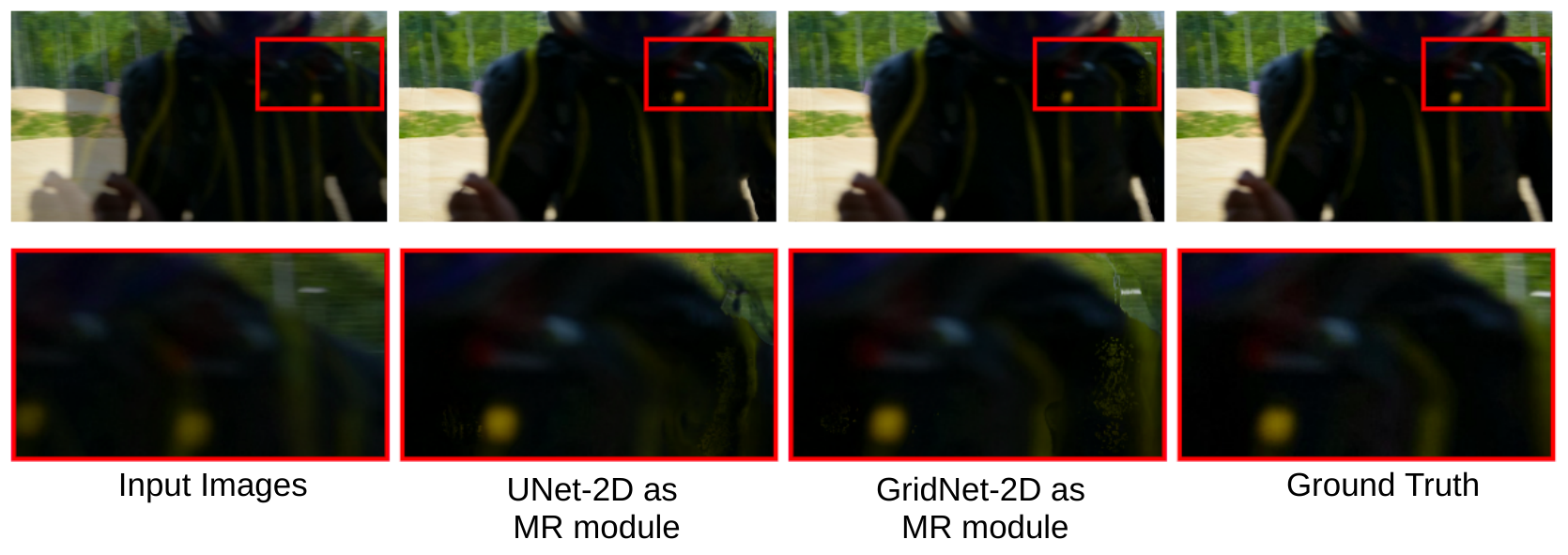}
    \vspace{-10px}
    \caption{Qualitative comparison between different MR modules.}
    \label{fig_ref}
    \vspace{-0.3cm}
\end{figure*}

\subsection{Objective Functions}
Following prior work \cite{superslomo}, we use following loss functions to train our model.

\textbf{Reconstruction Loss:} We use $L_1$ loss to capture the reconstruction quality of predicted intermediate frames.
Reconstruction loss $\mathcal{L}_r$ is given by,

\begin{equation}
    \mathcal{L}_r = \left\|\hat{{I}_{t}}-I_{t}\right\|_{1}
\end{equation}
Here, $\hat{{I}_{t}}$ and $I_{t}$ refer to the predicted interpolated RGB frame, ground-truth RGB frame respectively.

\textbf{Perceptual Loss:} Difference between features extracted from initial layers of a pre-trained image classification network can help to generate images of higher perceptual quality \cite{johnson2016perceptual}. 
Perceptual loss $\mathcal{L}_p$ is given by:
\begin{equation}
\mathcal{L}_p = \left\|\phi(\hat{{I}_{t}})-\phi(I_{t})\right\|_{2}
\end{equation}

where $\phi(.)$ denotes function that generates features from conv4\_3 layer of pretrained VGGNet-16.

\textbf{Warping Loss:} 
$L_1$ loss between warped frames and ground truth intermediate frames is used as warping loss because better flow predictions mean warped frames will be closer to ground truth intermediate frame.

\begin{equation}
    \mathcal{L}_w = \left\|I_t - bw(I_0, F^r_{t\rightarrow0})\right\|_1 + \left\|I_t - bw(I_1, F^r_{t\rightarrow1})\right\|_1 
\end{equation}

Here \textit{bw(.,.)} denotes the backward warping function.

\textbf{Smoothness Loss:} 
Total variation (TV) loss is used as smoothness loss to ensure smoothness in intermediate optical flow prediction.
\begin{equation}
    \mathcal{L}_s = \left\| \nabla F^r_{t\rightarrow0}\right\|_1 + 
    \left\| \nabla F^r_{t\rightarrow1}\right\|_1
\end{equation}

Our final loss is a linear combination of all the loss functions described above. 
\begin{equation}
    \mathcal{L} = \lambda_r \mathcal{L}_r + \lambda_p  \mathcal{L}_p + \lambda_w  \mathcal{L}_w + \lambda_s  \mathcal{L}_s
\end{equation}
We choose $\lambda_r = 204$, $\lambda_p = 0.005$, $\lambda_w = 102$ and $\lambda_s=1$ similar to unofficial SuperSloMo repository \cite{ssm_unoff}. When the model is trained with low learning rate at later phase, we turn off warping loss and smoothness loss by setting $\lambda_w$ and $\lambda_s$ to 0. This helps network to focus on improving the final reconstruction quality of interpolated frame.

\begin{table*}[!htb]
\centering
\caption{Quantitative comparison with state-of-the-art methods. Best and second best scores are colred in \textcolor{red}{{red}} and \textcolor{blue}{{blue}} respectively. * - TOFlow \cite{toflow} was trained on Vimeo-Triplet dataset, all other methods are trained in Vimeo-Septuplet dataset.}
\label{sota_comp}
\small
\begin{tabular}{|c|c|cc|cc|cc|cc|c|c|}
\hline
\multirow{2}{*}{Method}               & \multirow{2}{*}{\begin{tabular}[c]{@{}c@{}}Input\\ frames\end{tabular}} & \multicolumn{2}{c|}{Vimeo Septuplet}                                        & \multicolumn{2}{c|}{DAVIS}                                                  & \multicolumn{2}{c|}{HD}                                                    & \multicolumn{2}{c|}{GoPro}                                                   & \multirow{2}{*}{\begin{tabular}[c]{@{}c@{}}Params\\ (M)\end{tabular}} & \multirow{2}{*}{\begin{tabular}[c]{@{}c@{}}Runtime\\ (s)\end{tabular}} \\ \cline{3-10}
                                      &                                                                         & \multicolumn{1}{c|}{\textbf{PSNR}}             & \textbf{SSIM}              & \multicolumn{1}{c|}{\textbf{PSNR}}             & \textbf{SSIM}              & \multicolumn{1}{c|}{\textbf{PSNR}}            & \textbf{SSIM}              & \multicolumn{1}{c|}{\textbf{PSNR}}              & \textbf{SSIM}              &                                                                       &                                                                        \\ \hline
TOFlow* \cite{toflow}                 & 2                                                                       & \multicolumn{1}{c|}{33.46}                     & 0.9399                     & \multicolumn{1}{c|}{25.49}                     & 0.7577                     & \multicolumn{1}{c|}{\textcolor{blue}{30.94}}  & 0.8854                     & \multicolumn{1}{c|}{27.08}                      & 0.8286                     & \textcolor{red}{1.07}                                                                  & 0.10                                                                   \\ \hline
SepConv \cite{sepconv}                & 2                                                                       & \multicolumn{1}{c|}{33.04}                     & 0.9334                     & \multicolumn{1}{c|}{25.38}                     & 0.7428                     & \multicolumn{1}{c|}{30.24}                    & 0.8784                     & \multicolumn{1}{c|}{26.88}                      & 0.8166                     & 21.6                                                                  & \textcolor{blue}{0.024}                                                                  \\ \hline
SuperSloMo \cite{superslomo}          & 2                                                                       & \multicolumn{1}{c|}{33.46}                     & 0.9423                     & \multicolumn{1}{c|}{25.84}                     & 0.7765                     & \multicolumn{1}{c|}{30.37}                    & 0.8834                     & \multicolumn{1}{c|}{27.31}                      & 0.8367                     & 39.61                                                                 & 0.025                                                                  \\ \hline
CAIN \cite{cain}                      & 2                                                                       & \multicolumn{1}{c|}{31.70}                     & 0.9106                     & \multicolumn{1}{c|}{24.89}                     & 0.7235                     & \multicolumn{1}{c|}{29.22}                    & 0.8523                     & \multicolumn{1}{c|}{26.81}                      & 0.8076                     & 42.78                                                                 & \textcolor{red}{0.02}                                                                   \\ \hline
BMBC\footnotemark \cite{park2020bmbc} & 2                                                                       & \multicolumn{1}{c|}{31.34}                     & 0.9054                     & \multicolumn{1}{c|}{23.50}                     & 0.6697                     & \multicolumn{1}{c|}{-}                        & -                          & \multicolumn{1}{c|}{24.62}                      & 0.7399                     & \textcolor{blue}{11.0}                                                                  & 0.41                                                                   \\ \hline
Tridirectional \cite{tridirectional}  & 3                                                                       & \multicolumn{1}{c|}{32.73}                     & 0.9331                     & \multicolumn{1}{c|}{25.24}                     & 0.7476                     & \multicolumn{1}{c|}{29.84}                    & 0.8692                     & \multicolumn{1}{c|}{26.80}                      & 0.8180                     & 10.40                                                                 & 0.19                                                                   \\ \hline
QVI \cite{xu2019quadratic}            & 4                                                                       & \multicolumn{1}{c|}{\textcolor{blue}{{34.50}}} & \textcolor{blue}{{0.9521}} & \multicolumn{1}{c|}{\textcolor{blue}{{27.36}}} & \textcolor{red}{{0.8298}}  & \multicolumn{1}{c|}{30.92}                    & \textcolor{blue}{{0.8971}} & \multicolumn{1}{c|}{\textcolor{blue}{{28.80} }} & \textcolor{blue}{{0.8781}} & 29.22                                                                 & 0.10                                                                   \\ \hline
FLAVR  \cite{kalluri2020flavr}        & 4                                                                       & \multicolumn{1}{c|}{33.56}                     & 0.9372                     & \multicolumn{1}{c|}{25.74}                     & 0.7589                     & \multicolumn{1}{c|}{29.96}                    & 0.8758                     & \multicolumn{1}{c|}{27.76}                      & 0.8436                     & 42.06                                                                 & 0.20                                                                   \\ \hline
Ours                                  & 4                                                                       & \multicolumn{1}{c|}{\textcolor{red}{{34.99}}}  & \textcolor{red}{{0.9544}}  & \multicolumn{1}{c|}{\textcolor{red}{{27.53}}}  & \textcolor{blue}{{0.8281}} & \multicolumn{1}{c|}{\textcolor{red}{{31.49}}} & \textcolor{red}{{0.9000}}  & \multicolumn{1}{c|}{\textcolor{red}{{29.08}}}   & \textcolor{red}{{0.8826}}  & 20.92                                                                    & 0.32                                                                      \\ \hline
\end{tabular}
\end{table*}
\footnotetext{BMBC encountered out-of-memory error when tested on HD dataset.}

\begin{figure*}[!ht]
    \centering
    \includegraphics[width=0.85\textwidth]{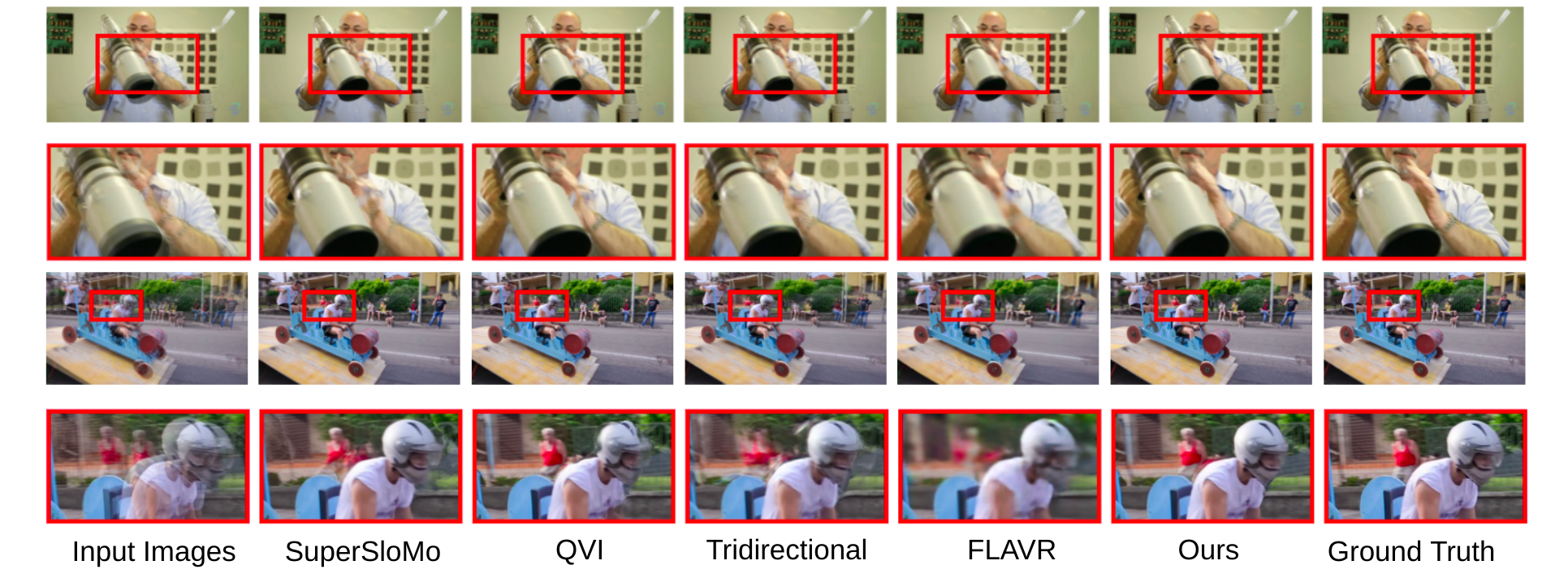}
    \vspace{-10px}
    \caption{Qualitative comparison of our method with other state-of-the-art algorithms.}
    \vspace{-0.5cm}
    \label{sota_comp_fig1}
\end{figure*}

\subsection{Experiments on model configurations}
\label{modelconfig}
In this section, we perform comparative studies among the different choices available for NME (UNet-2D \cite{superslomo}, UNet-3D \cite{kalluri2020flavr}, GridNet-3D) and MR (UNet-2D \cite{unet}, GridNet-2D \cite{gridnet}) modules to determine the best performing configuration.

\textbf{Choice of NME module:} We experiment with three different architectures for NME module: 1) UNet-2D \cite{superslomo}, 2) UNet-3D  \cite{kalluri2020flavr}, and 3) novel GridNet-3D proposed in this paper. We illustrate the quantitative performance with different NME modules in Table \ref{nme_abl_tab} along with number of parameters and runtimes. We observe that 3D-CNN version of NME modules perform superior to UNet-2D in general. Further, GridNet-3D performs better than UNet-3D in DAVIS, HD and GoPro datasets while having less parameters and runtime.

\textbf{Choice of MR modules:} We experiment with two types of motion refinement modules: UNet-2D \cite{unet} and GridNet-2D \cite{gridnet}. We use a standard encoder-decoder architecture with skip connections for UNet-2D. In GridNet-2D, encoder and decoder blocks are laid out in a grid-like fashion to carry through multi-scale feature maps till the final layer. Quantitative comparison in Table \ref{flow_ref} shows that using GridNet-2D as MR module performs significantly better than UNet-2D. Qualitative comparison in Figure \ref{fig_ref} illustrates that GridNet-2D reduces the smudge effect in interpolated frame compared to UNet-2D. From Table \ref{flow_ref}, we can also infer that using GridNet-2D as MR module reduces total number of parameters of the model while the runtime remains constant.

Based on these experiments, we use GridNet-3D in NME module and GridNet-2D as MR module in state-of-the-art comparisons and in ablation studies unless specified otherwise.

\subsection{Comparison with state-of-the-arts \label{sec:sota_comparison}}
We compare our model with multiple state-of-the-art methods: TOFlow \cite{toflow}, Sepconv-$\mathcal{L}_1$ \cite{sepconv}, SuperSloMo \cite{superslomo}, CAIN \cite{cain}, BMBC \cite{park2020bmbc}, QVI \cite{xu2019quadratic}, Tridirectional \cite{tridirectional} and FLAVR \cite{kalluri2020flavr}. For comparison with TOFlow, official pretrained model \cite{toflow_repo} is used. For all other models, we train them on Vimeo-Septuplet train set with same learning rate schedule and batch size as ours for fair comparison. We use unofficial repositories of SuperSloMo \cite{ssm_unoff} and Sepconv \cite{sepconv_unoff} to train the corresponding models. Please note, official pretrained models of other methods might produce different results due to difference in training data and training settings.
During evaluation, Peak Signal-to-Noise ratio (PSNR) and Structural Similarity (SSIM) \cite{ssim} are used as evaluation metric to compare performances.
Quantitative comparisons with state-of-the-art methods on Vimeo, DAVIS, HD and GoPro datasets are shown in Table \ref{sota_comp}. Number of parameters and average runtime to produce a frame of resolution $256 \times 448$ on NVIDIA 1080Ti GPU for each model is also reported.

Our method achieves best PSNR and SSIM scores in Vimeo, HD and GoPro datasets. Our method performs best in PSNR and second best in SSIM metric on DAVIS dataset. 
Qualitative comparison with other methods is shown in Figure \ref{sota_comp_fig1}.

\begin{table*}[!ht]
\vspace{-0.3cm}
\caption{Effect of different input features to 3D CNN.}
\vspace{-0.3cm}
\centering
\small
\begin{tabular}{|c|cc|cc|cc|cc|c|c|}
\hline
\multirow{2}{*}{Input} & \multicolumn{2}{c|}{Vimeo Septuplet}                  & \multicolumn{2}{c|}{DAVIS}                            & \multicolumn{2}{c|}{HD}                               & \multicolumn{2}{c|}{GoPro}                            & \multirow{2}{*}{\begin{tabular}[c]{@{}c@{}}Params\\ (M)\end{tabular}} & \multirow{2}{*}{\begin{tabular}[c]{@{}c@{}}Runtime\\ (s)\end{tabular}} \\ \cline{2-9}
                       & \multicolumn{1}{c|}{\textbf{PSNR}}  & \textbf{SSIM}   & \multicolumn{1}{c|}{\textbf{PSNR}}  & \textbf{SSIM}   & \multicolumn{1}{c|}{\textbf{PSNR}}  & \textbf{SSIM}   & \multicolumn{1}{c|}{\textbf{PSNR}}  & \textbf{SSIM}   &                                                                       &                                                                        \\ \hline
RGB                    & \multicolumn{1}{c|}{34.12}          & 0.9474          & \multicolumn{1}{c|}{26.34}          & 0.7883          & \multicolumn{1}{c|}{30.80}          & 0.8854          & \multicolumn{1}{c|}{28.34}          & 0.8642          & \textbf{61.89}                                                                 & \textbf{0.23}                                                                   \\ \hline
Flow + Occlusion       & \multicolumn{1}{c|}{\textbf{34.70}} & \textbf{0.9532} & \multicolumn{1}{c|}{\textbf{27.32}} & \textbf{0.8260} & \multicolumn{1}{c|}{\textbf{31.02}} & \textbf{0.8944} & \multicolumn{1}{c|}{\textbf{28.81}} & \textbf{0.8798} & 78.11                                                                 & 0.37                                                                   \\ \hline
\end{tabular}
\label{input_abl}
\vspace{-0.3cm}
\end{table*}

\begin{figure*}[!ht]
    \centering
    \includegraphics[width=0.65\textwidth]{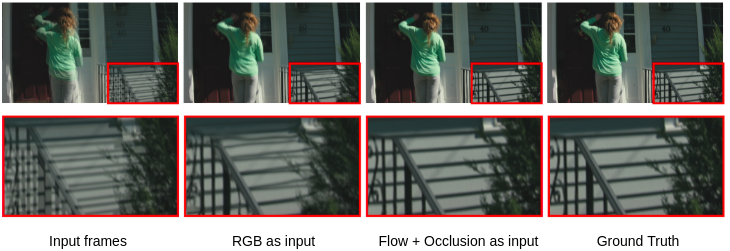}
    \vspace{-10px}
    \caption{Qualitative comparison between RGB and Flow+Occlusion as input to 3D CNN.}
    \label{rgb_fig}
    \vspace{-0.1cm}
\end{figure*}

\begin{table*}[!ht]
\caption{Quantitative significance of BFE, MR and BME modules.}
\vspace{-0.3cm}
\centering
\small
\begin{tabular}{|c|cc|cc|cc|cc|c|c|}
\hline
\multirow{2}{*}{}       & \multicolumn{2}{c|}{Vimeo Septuplet}                  & \multicolumn{2}{c|}{DAVIS}                            & \multicolumn{2}{c|}{HD}                               & \multicolumn{2}{c|}{GoPro}                            & \multirow{2}{*}{\begin{tabular}[c]{@{}c@{}}Params\\ (M)\end{tabular}} & \multirow{2}{*}{\begin{tabular}[c]{@{}c@{}}Runtime\\ (s)\end{tabular}} \\ \cline{2-9}
                        & \multicolumn{1}{c|}{\textbf{PSNR}}  & \textbf{SSIM}   & \multicolumn{1}{c|}{\textbf{PSNR}}  & \textbf{SSIM}   & \multicolumn{1}{c|}{\textbf{PSNR}}  & \textbf{SSIM}   & \multicolumn{1}{c|}{\textbf{PSNR}}  & \textbf{SSIM}   &                                                                       &                                                                        \\ \hline
without BFE, MR and BME & \multicolumn{1}{c|}{33.91}          & 0.9443          & \multicolumn{1}{c|}{26.05}          & 0.7686          & \multicolumn{1}{c|}{30.72}          & 0.8811          & \multicolumn{1}{c|}{28.12}          & 0.8583          & \textbf{42.07}                                                                 & \textbf{0.20}                                                                   \\ \hline
with BFE, MR and BME    & \multicolumn{1}{c|}{\textbf{34.12}} & \textbf{0.9474} & \multicolumn{1}{c|}{\textbf{26.34}} & \textbf{0.7883} & \multicolumn{1}{c|}{\textbf{30.80}} & \textbf{0.8854} & \multicolumn{1}{c|}{\textbf{28.34}} & \textbf{0.8642} & 61.89                                                                 & 0.23                                                                   \\ \hline
\end{tabular}
\label{rev_abl_tab}
\vspace{-0.3cm}
\centering
\end{table*}

\begin{figure*}[!ht]
    \centering
    \includegraphics[width=0.65\textwidth]{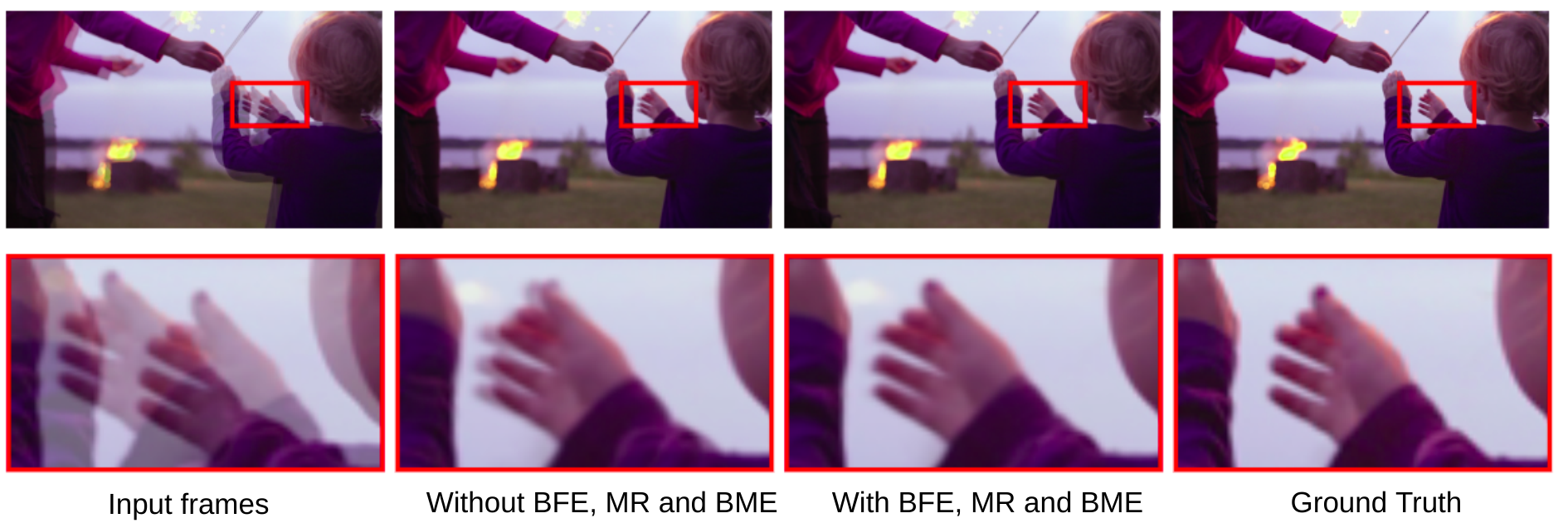}
    \vspace{-10px}
    \caption{Qualitative comparison between intermediate flowmap and blending mask estimation with and without BFE, MR and BME modules.}
    \label{rev_fig}
    \vspace{-0.5cm}
\end{figure*}

\begin{figure*}[!ht]
    \centering
    \includegraphics[width=0.8\textwidth]{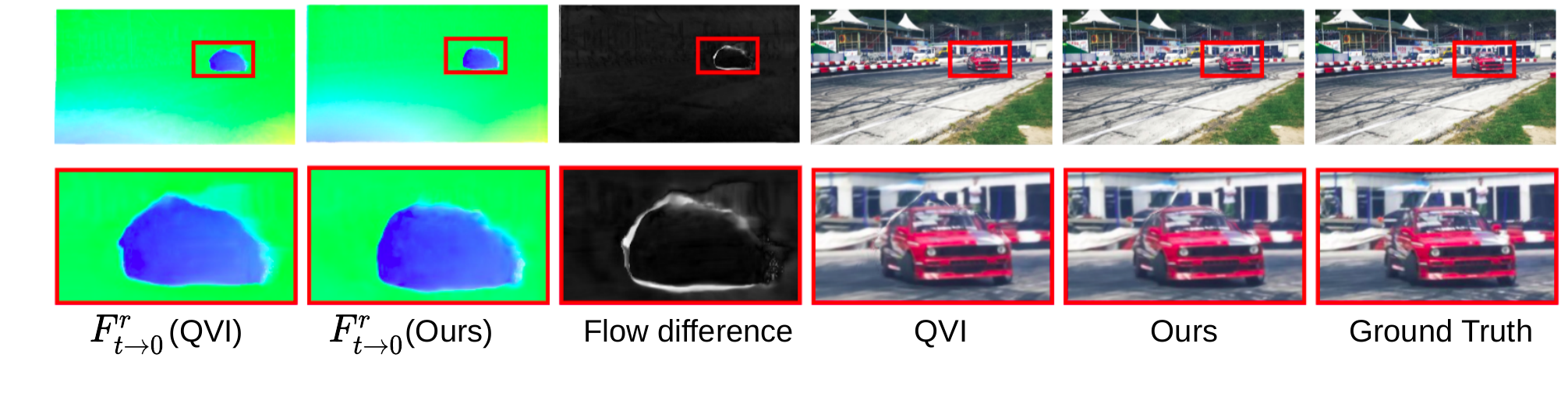}
    \vspace{-0.8cm}
    \caption{Intermediate flow visualization between QVI and our approach.}
    \vspace{-0.5cm}
    \label{flovis_abl}
\end{figure*}

\subsection{Ablation studies}
\label{ablation_sec}
\textbf{Choice of input features (RGB vs. Flow+Occlusion):} To demonstrate the importance of flow and occlusion maps, we perform an experiment where we use RGB frames as input to the 3D CNN. Quantitative comparison between these two approaches are shown in Table \ref{input_abl} with number of parameters and runtimes. Both experiments in Table \ref{input_abl} use UNet-2D as MR module. We observe that Flow+Occlusion maps as input performs better than RGB frames. Qualitative comparison in Figure \ref{rgb_fig} shows that interpolated results are more accurate when Flow+Occlusion maps are used compared to RGB.
Note that, our model with RGB input already performs better than FLAVR \cite{kalluri2020flavr} (refer to Table \ref{sota_comp}). This signifies that frame generation by hallucinating pixels from scratch \cite{kalluri2020flavr} is hard for neural networks to achieve than frame generation by warping neighborhood frames.

\textbf{Importance of BFE, MR and BME modules: } 
To understand the importance of BFE, MR and BME modules, we re-purpose the NME module to directly predict non-linear backward flows $F_{t\rightarrow0}$, $F_{t\rightarrow1}$ and blending mask $M$. In this experiment, we use RGB frames as input to NME module. The quantitative comparison in Table \ref{rev_abl_tab} illustrates that the direct estimation of backward flows, mask (without BFE, MR and BME) performs sub-par to estimating them with BFE, MR (UNet-2D) and BME modules. Added to this, qualitative comparisons in Figure \ref{rev_fig} shows that the direct estimation of backward flows may lead to ghosting artifacts due to inaccurate flow estimation near motion boundaries.

\textbf{Intermediate flow visualizations:} We visualize the backward flow $F^r_{t\rightarrow0}$ estimated by QVI \cite{xu2019quadratic} and our approach in Figure \ref{flovis_abl}. We notice that erroneous results in QVI's \cite{xu2019quadratic} interpolated frame is caused by incorrect estimation of the backward flow. However, our method remedies this by accurately estimating the backward flow as visualized in the absolute flow difference map in Figure \ref{flovis_abl}.

\vspace{-0.3cm}
\section{Conclusion}

\label{sec:conclusion}
In this paper, we presented a 3D CNN based frame interpolation algorithm in which the bi-directional flow and occlusion maps between neighboring frames are passed as input to a 3D CNN to predict per-pixel non-linear motion. 
This makes our network flexible to choose between linear and quadratic motion models instead of a fixed motion model as used in prior work. Our method achieves state-of-the-art results in multiple datasets. 
Since flow and occlusion estimates from \textit{PWCNet-Bi-Occ} are often not accurate and hence can create a performance bottleneck in interpolation task, further research can explore whether inclusion of RGB frames as input to 3D CNN can improve the performance. Finally, flow representation estimation for cubic modeling can also be investigated in future.


{\small
\bibliographystyle{ieee_fullname}
\bibliography{egbib}
}

\onecolumn

\section{Appendix}

\subsection{Importance of using four frames}
In order to show the effectiveness of using four frames in our network, we report results of our model using only two frames ($I_0,I_1$) as input. Since our model expects four frames as input, we use frame repetition ($I_0,I_0,I_1,I_1$) in this experiment. Quantitative comparison in Table-\ref{two_frames} shows that our model indeed benefits from using four frames as input.

\begin{table*}[!htp]
\caption{Quantitative comparison between using two frames and four frames}
\label{two_frames}
\centering
\begin{tabular}{|c|cc|cc|cc|cc|}
\hline
\multirow{2}{*}{\begin{tabular}[c]{@{}c@{}}No. of\\ input frames\end{tabular}} & \multicolumn{2}{c|}{Vimeo Septuplet} & \multicolumn{2}{c|}{DAVIS}          & \multicolumn{2}{c|}{HD}             & \multicolumn{2}{c|}{GoPro}          \\ \cline{2-9} 
                                                                               & \multicolumn{1}{c|}{PSNR}   & SSIM   & \multicolumn{1}{c|}{PSNR}  & SSIM   & \multicolumn{1}{c|}{PSNR}  & SSIM   & \multicolumn{1}{c|}{PSNR}  & SSIM   \\ \hline
2                                                                              & \multicolumn{1}{c|}{33.61}  & 0.9438 & \multicolumn{1}{c|}{26.04} & 0.7737 & \multicolumn{1}{c|}{30.97} & 0.8901 & \multicolumn{1}{c|}{27.27}      &   0.8347     \\ \hline
4                                                                              & \multicolumn{1}{c|}{\textbf{34.99}}  & \textbf{0.9544} & \multicolumn{1}{c|}{\textbf{27.53}} & \textbf{0.8281} & \multicolumn{1}{c|}{\textbf{31.49}} & \textbf{0.9000} & \multicolumn{1}{c|}{\textbf{29.08}} & \textbf{0.8826} \\ \hline
\end{tabular}
\end{table*}

\subsection{Multi-frame interpolation}

We have tested 4x interpolation results (generating 3 intermediate frames) in a recursive way on GoPro test set and compared it with QVI \cite{xu2019quadratic}. Quantitative results are shown in Table-\ref{multi_frame}. We can see that our model can perform better than QVI \cite{xu2019quadratic} on multi-frame interpolation case too. 

\begin{table}[!h]
\caption{Multi-frame interpolation results on GoPro dataset.}
\centering
\begin{tabular}{|c|c|c|}
\hline
Method & PSNR  & SSIM   \\ \hline
QVI    & 29.36 & 0.8964 \\ \hline
Ours   & \textbf{29.86} &\textbf{ 0.9021} \\ \hline
\end{tabular}
\label{multi_frame}
\end{table}

\subsection{Parameter and runtime analysis of different components}

In Table-\ref{indiv_analysis}, we have reported number of parameters and average runtime of different components of our network.

\begin{table}[!htp]
\caption{Component-wise parameter and runtime analysis}
\label{indiv_analysis}
\centering
\begin{tabular}{|c|c|c|c|}
\hline
\begin{tabular}[c]{@{}c@{}}Component\\ name\end{tabular}                & Specification & \begin{tabular}[c]{@{}c@{}}Params\\ (M)\end{tabular} & \begin{tabular}[c]{@{}c@{}}Runtime\\ (s)\end{tabular} \\ \hline
\begin{tabular}[c]{@{}c@{}}Flow and occlusion\\  estimator\end{tabular} & -             & 16.19                                                & 0.12                                                  \\ \hline
3D CNN                                                                  & UNet-3D       & 42.06                                                & 0.20                                                  \\ \hline
3D CNN                                                                  & GridNet-3D    & 2.44                                                 & 0.15                                                  \\ \hline
BFE                                                                     & -             & 0                                                    & 0.02                                                  \\ \hline
MR                                                                      & UNet          & 19.81                                                & 0.016                                                 \\ \hline
MR                                                                      & GridNet       & 2.25                                                 & 0.021                                                 \\ \hline
BME                                                                     & -             &    0.04                                                  & 0.002                                                 \\ \hline
Frame Synthesis                                                         & -             & 0                                                    & 0.002                                                 \\ \hline
\end{tabular}
\end{table}

\end{document}